\documentclass[letterpaper, 10 pt, journal, twoside]{IEEEtran}
\IEEEoverridecommandlockouts                              

\usepackage{pifont}
\newcommand{\xmark}{\ding{53}}%
\usepackage{graphics}
\usepackage[pdftex]{graphicx}
\usepackage[tight,footnotesize]{subfigure}
\usepackage{multirow}
\usepackage[cmex10]{amsmath}
\usepackage{amssymb}
\usepackage{amsfonts}
\usepackage{algpseudocode}
\usepackage{algorithm}
\usepackage{booktabs}
\usepackage{textcomp}
\usepackage{url}
\usepackage{color}
\usepackage{siunitx}
\usepackage{soul}

\usepackage{threeparttable} 
\usepackage{threeparttablex} 
\usepackage{wrapfig}
\usepackage{babel}
\usepackage[small]{caption}
\usepackage{placeins}
\usepackage{float}
\usepackage{algorithm}
\usepackage{algpseudocode}
\usepackage{makecell}

\usepackage{footnote}
\usepackage{tabularray}
\usepackage{stfloats}

\usepackage{multirow}
\usepackage{wrapfig}
\usepackage{colortbl}
\usepackage[normalem]{ulem}

\newcolumntype{g}{>{\columncolor{CuGray}}c}
\newcolumntype{z}{>{\columncolor{CuGray}}l}

\renewcommand{\paragraph}[1]{\noindent\textbf{#1.}\,\,}

\usepackage[table]{xcolor}
\newcommand{\cse}{\texttt{CSE}}

\newcommand{\office}{\emph{Office}}
\newcommand{\hospital}{\emph{Hospital}}
\newcommand{\warehouse}{\emph{Warehouse}}

\usepackage{xspace}

\def\onedot{.\@\xspace}
\def\eg{\emph{e.g}\onedot} 
\def\ie{\emph{i.e}\onedot} 
 
\def\etc{\emph{etc}\onedot} 
 
\def\etal{\emph{et al}\onedot}

\newcommand{\Fref}[1]{Fig.~\ref{#1}}
\newcommand{\Tref}[1]{Table~\ref{#1}}

\newcommand{\be}{\begin{eqnarray}}
\newcommand{\ee}{\end{eqnarray}}
\newcommand{\bee}{\begin{eqnarray*}}
\newcommand{\eee}{\end{eqnarray*}}

\newcommand{\matrixb}{\left[ \begin{array}}
\newcommand{\matrixe}{\end{array} \right]}

\title{
A Benchmark Dataset for Collaborative SLAM in Service Environments
}
\author{Harin Park$^{1}$,  Inha Lee$^{1}$,  Minje Kim$^{1}$, Hyungyu Park$^{1}$ and Kyungdon Joo$^{2,\dag}$ \\
\thanks{
Manuscript received: May 29, 2024; Revised September 11, 2024; Accepted October 11, 2024.
This paper was recommended for publication by Editor Cesar Cadena Lerma upon evaluation of the Associate Editor and Reviewers' comments.
This work was supported by Institute of Information \& communications Technology Planning \& Evaluation (IITP) grant funded by the Korea government (MSIT) (No.RS-2022-II220907, Development of AI Bots Collaboration Platform and Self-organizing AI, No.RS-2020-II201336, Artificial Intelligence Graduate School Program (UNIST)) and the National Research Foundation of Korea (NRF) grant funded by the Korea government (MSIT) (No.RS-2024-00457065).
}
\thanks{
$^{1}$Harin Park, Inha Lee,  Minje Kim and Hyungyu Park are with the Artificial Intelligence Graduate School, UNIST, South Korea. 
{(\tt\footnotesize \{harinp33, epsilon8854, minje617, hyungyu\}@unist.ac.kr)}
}
\thanks{
$^{2}$Kyungdon Joo is with the Artificial Intelligence Graduate School and the Department of Computer Science and Engineering, UNIST, Ulsan, South Korea. {(\tt\footnotesize kdjoo369@gmail.com, kyungdon@unist.ac.kr})}
\thanks{$^{\dag}$Corresponding author}
\thanks{Digital Object Identifier (DOI): see top of this page.}
} 

\begin{document}
\maketitle

\begin{abstract}
    We introduce a new multi-modal collaborative SLAM (C-SLAM) dataset for multiple service robots in various indoor service environments, called \texttt{C}-SLAM dataset in \texttt{S}ervice \texttt{E}nvironments (\cse).
    We use the NVIDIA Isaac Sim to generate data in various indoor service environments with the challenges that may occur in real-world service environments.
    By using the simulator, we can provide precisely time-synchronized sensor data, such as stereo RGB/depth, IMU, and ground truth (GT) poses.
    We configure three common indoor service environments (\hospital, \office, and \warehouse), each featuring dynamic objects performing motions suited to the environment.
    In addition, we drive the robots to mimic the actions of real service robots.
    Through these factors, we generate a realistic C-SLAM dataset for multiple service robots.
    We demonstrate our \cse~dataset by evaluating diverse state-of-the-art single-robot SLAM and multi-robot SLAM methods.
    Additionally, we provide a detailed tutorial on generating C-SLAM data using the simulator.
    Our tutorial and dataset are available at \texttt{\small\url{https://github.com/vision3d-lab/CSE_Dataset}}.
\end{abstract}

\begin{figure*}[t]
\centering
\includegraphics[width=0.85\linewidth]{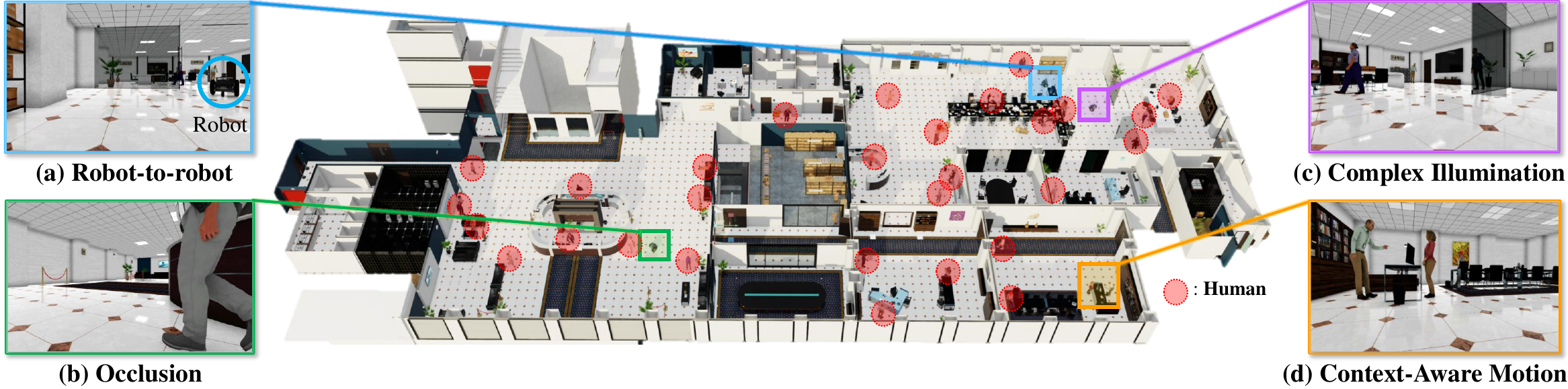}
\caption{
\textbf{Illustration of the \cse~dataset in \emph{Office} environments}.
        The \cse~dataset is obtained from realistic service environments, including multiple dynamic objects, indicated by a red circle. Our environments contain diverse characteristics, and each box shows the features as seen from the robot camera view.
        (a) Robot-to-robot interaction (\textsf{\footnotesize Follow}). The blue circle is the robot driving in front of it.
        (b) Occlusion from dynamic objects. 
        (c) Complex illumination due to reflective material. 
        (d) Dynamic objects performing motions suitable for the environment.
    }
    \vspace{-3mm}
\label{fig:overview}
\end{figure*}



\section{Introduction}  \label{sec:intro}

\IEEEPARstart{I}{ntelligent} agents, such as personal robots and autonomous vehicles, have increasingly become a part of our daily lives and play significant roles.
In particular, service robots have started to replace simple yet human resources-required tasks, such as serving, path guidance, cleaning, delivery, \etc~\cite{ouyang2021collaborative,paluch2020service}.
To this end, service robots are required to understand unknown environments~\cite{paluch2020service}, where simultaneous localization and mapping (SLAM), which estimates the pose of the robot itself and builds a map of an unknown environment simultaneously, is one of the most fundamental techniques for service robots~\cite{openloris}.
Through SLAM, service robots can perceive their surroundings and perform their tasks.

Specifically, service environments, where robots operate and may interact with people~\cite{lo2017iterative}, have become diverse~(\eg, from static spaces to complex indoor or outdoor environments) and have begun to require more complex tasks that are difficult for a single agent to handle.
These changes naturally have led to an interest in multiple agents from single agents.
Furthermore, SLAM algorithms, the basis of robot perception, have also begun promoting performance improvement through collaboration between multiple agents. 
Accordingly, a new SLAM task, \emph{Collaborative SLAM} (\emph{C-SLAM} in short) for multiple agents, has been developed in the robotics community~\cite{jennings1999cooperative, fox2000probabilistic} and aims to improve the robustness, and accuracy of localization and mapping by exchanging spatial information among multiple agents~\cite{ouyang2021collaborative}.

Despite this progress, C-SLAM remains limited in terms of benchmarking~\cite{towards_coslam_survey}.
While standardized benchmarks for SLAM for a single robot have emerged extensively, systematic evaluation techniques and datasets for C-SLAM are still lacking~\cite{towards_coslam_survey}.
For example, the OpenLORIS-Scene dataset~\cite{openloris}, as a SLAM dataset for a single service robot, encompasses various challenges encountered in service environments, such as textureless scenes, dynamic objects, and viewpoint changes.
However, early datasets for C-SLAM~\cite{UTIAS,airmuseum} are acquired only in static and indoor experimental environments, excluding dynamic objects.
To alleviate these {gaps}, a few datasets that include dynamic objects, such as humans, have recently been proposed~\cite{s3e,ford,graco,tian2023resilient}, but they are mainly acquired from urban outdoor scenes or limited indoor environments (\eg, only a small corridor or a room-size laboratory). 
In other words, there is still a lack of diversity in service environments for multi-robot in terms of benchmark datasets~(see \Tref{Intro:comparison}).

In the case of service robots, they operate for long periods of time in various indoor environments, where they have diverse interactions.
Concretely, the service robots navigate complex indoor spaces, collaborate with others, or interact with people in service environments, such as hospital, restaurant, and office~\cite{ouyang2021collaborative,paluch2020service}.
Notably, while multiple service robots are in operation, challenging scenarios arise for performing C-SLAM.
For example, the robots may encounter homogeneous scenes or severe occlusions/large rotations caused by dynamic objects.
Motivated by this fact, in this work, we introduce a new multi-modal C-SLAM dataset for multiple service robots in indoor service environments, called \cse~dataset; especially, our dataset includes various indoor service environments, such as \hospital, \office, and \warehouse~(see \emph{Office} in \Fref{fig:overview}), and mimics diverse and challenging scenarios for service robots.

In constructing our dataset, we consider several essential factors that must be satisfied as a C-SLAM dataset for service robots.
1)~Each robot must provide time-synchronized and abundant sensor modalities that can allow us to demonstrate SLAM for different sensor combinations.
2)~Multi-robot should explore diverse paths in various service environments, and precisely time-synchronized GT poses for each robot are essential for evaluating C-SLAM. 
It should be noted that acquiring accurate time-synchronized sensor data and GT poses among multi-robot is non-trivial in the real world.
3)~Multi-robot must reproduce various scenarios that can occur in real service environments, such as avoiding dynamic objects during path planning.
To satisfy the above factors, we propose a new synthetic C-SLAM dataset for multiple service robots using a simulator, NVIDIA Isaac Sim~\cite{isaac_sim}, {which} provides photo-realistic sensor data.
Based on the NVIDIA Isaac Sim, we acquire time-synchronized sensor data and accurate GT poses. 
We also configure various service environments with the challenges that arise in real service environments.
Each environment is separated into static and dynamic, based on the on/off of dynamic objects.
In addition, we build a realistic dataset by simulating real service robot actions.
The main characteristics of our datasets are as follows:
\begin{itemize}
    \item We propose a new synthetic C-SLAM dataset for multiple service robots.
    Each robot includes precisely time-synchronized stereo RGB, stereo depth, inertial measurement unit (IMU), and GT pose.
    \item We acquire data in diverse service environments: \hospital, \warehouse, \office.
    For each environment, we place the dynamic objects {with} suitable actions and include diverse challenging cases.
    \item We construct the scenarios considering intra/inter-robot loop closures to facilitate the appropriate evaluation of C-SLAM. 
    In addition, we separate each environment into static and dynamic to acquire data in the same scenario. This allows us to evaluate the efficiencies of SLAM algorithms dealing with dynamic objects.
    \item We validate our dataset by evaluating diverse state-of-the-art single-robot SLAM and multi-robot SLAM. 
\end{itemize}



\section{Related Work}  \label{sec:related_work}

\subsection{SLAM Datasets for Single-Robot}
Various datasets for {a} single robot that incorporates diverse environments and sensor modalities have been proposed in the literature.
In particular, there are several public SLAM datasets that are widely used for single-robot~\cite{euroc,kitti,tum_rgbd,tartanair}.
Among them, KITTI~\cite{kitti}, EuRoC-MAV~\cite{euroc}, and TUM RGB-D~\cite{tum_rgbd} are well-known public benchmark datasets for single-robot SLAM.
However, they are acquired in limited scenarios and do not reflect the complexity of real-world {environments}~\cite{tartanair}.
{Furthermore}, they lack diversity in environments and conditions, such as weather and lighting.

To address these limitations, Wang~\etal~\cite{tartanair} propose a synthetic SLAM dataset, TartanAir, which considers challenging conditions in various environments.
They build data focusing on challenging environments with diverse motion patterns, adverse weather, and dynamic objects.
By using simulations, they provide accurate sensor data, such as stereo RGB image, depth image, camera poses, \etc
In contrast, the OpenLORIS-Scene~\cite{openloris} is a lifelong SLAM dataset acquired in real service environments with diverse challenges. 
It includes elements found in real service environments (\eg, dynamic objects and textureless scenes) and collects data from offices and markets where service robots operate.

Unfortunately, despite advances in datasets for {a} single robot, they still have limitations in handling complex situations and performing multiple tasks simultaneously.

\begin{table*}[]
    \centering
    \caption{\textbf{Comparison of existing C-SLAM datasets for multi-robot.} Time synchronization columns are referenced by~\cite{s3e}.}
    \label{Intro:comparison}
    \begin{threeparttable}
    \resizebox{0.95\linewidth}{!}{%
    \begin{tabular}{c|cccc|c|cc|c|cc}  \toprule 
            \multirow{2}{*}{Dataset} & \multicolumn{4}{c|}{Sensors} & \multirow{2}{*}{Platforms} & \multicolumn{2}{c|}{Environment} & \multirow{2}{*}{Ground Truth Pose} & \multicolumn{2}{c}{Time sync}  \\
            & RGB & Depth & IMU & LiDAR &  & Type & Static / Dynamic &  & Intra & Inter \\
            \midrule
            UTIAS~\cite{UTIAS}     & \checkmark\tnote{$\dagger$} &  &  &  & UGV & Indoor  & S & Motion capture & Sw & NTP \\
            AirMuseum~\cite{airmuseum}        & \checkmark\tnote{$\dagger$} &  & \checkmark &  & UGV, UAV & Indoor  & S & SfM & Sw & NTP \\
            Ford-AV~\cite{ford}      &\checkmark &  & \checkmark & \checkmark & Vehicle & Outdoor  & D & GPS-IMU, SLAM corrected & -- & GNSS \\
            S3E~\cite{s3e}          & \checkmark &  & \checkmark & \checkmark & UGV & Outdoor, Indoor  & S & RTK, Motion capture & Hw & GNSS, PTPv2 \\
            GRACO~\cite{graco}          & \checkmark & & \checkmark & \checkmark & UGV, UAV & Outdoor  & D & GNSS\,/\,INS & Hw & GNSS \\
            Tian \textit{et al.}~\cite{tian2023resilient}        & \checkmark & \checkmark & \checkmark & \checkmark & UGV, UAV & Outdoor, Indoor  & D & Point-cloud by LiDAR, GPS & -- & NTP \\
            {SubT-MRS}~\cite{subtmrs}  & \checkmark & \checkmark & \checkmark & \checkmark & UGV, UAV, Legged & Outdoor, Indoor & S & 3D Scanner & Hw & -- \\
            {CERBERUS}~\cite{cerberus}  & \checkmark &  & \checkmark & \checkmark & UGV, UAV, Legged & Outdoor, Indoor & S & DARPA-provided map, point cloud  & Sw & PTP \\
            \midrule
            \textbf{Ours}      & \checkmark & \checkmark & \checkmark & \checkmark\tnote{*} & UGV & \textbf{\cellcolor{gray!15}{Service Env. (Indoor)}} & \textbf{\cellcolor{gray!15}{S + D}} & Simulator (NVIDIA Isaac Sim) & Simulator & Simulator \\
            \bottomrule
        \end{tabular}
        }
        \begin{scriptsize}
        \begin{tablenotes}
            \item $\dagger$ Only monocular RGB modality. 
            \item * We provide a function that converts depth images to pseudo-LiDAR data.
        \end{tablenotes}
        \end{scriptsize}
    \end{threeparttable}
    \vspace{-3mm}
\end{table*}

\subsection{C-SLAM Datasets for Multi-Robot}
The goal of C-SLAM is to enhance the efficiency and accuracy that surpass the capabilities of single-robot SLAM by integrating data from each robot to form globally consistent maps and state estimates~\cite{towards_coslam_survey}.
Due to their advantages, such as enabling mapping over large areas and facilitating efficient task execution, C-SLAM has gained significant attention in research. 
Nevertheless, there is still a lack of benchmark datasets for evaluating and developing C-SLAM.

UTIAS~\cite{UTIAS} is the first dataset for multi-robot SLAM. 
UTIAS presents a 2D multi-robot SLAM dataset based on 15 distinct landmarks in a 15m$\times$8m indoor environment, using five robots equipped with a monocular camera. 
However, it is limited to a constrained indoor experimental space and offers only 2D GT poses.
AirMuseum~\cite{airmuseum} and GRACO~\cite{graco} are multi-robot SLAM datasets utilizing heterogeneous agent platforms.
AirMuseum is collected in an indoor environment with multiple ground robots and drones equipped with Apriltag markers, providing GT trajectories through Structure from Motion (SfM).
But it assumes a static indoor environment.
GRACO involves the use of ground robots and drones in outdoor urban scenes.
However, they do not include challenges like occlusions from dynamic objects that robots may face in the real world.
Additionally, S3E~\cite{s3e} proposes a long-term multi-modal dataset using multiple ground robots in both indoor and outdoor environments based on four well-designed trajectory paradigms.
Tian~\etal~\cite{tian2023resilient} acquire data using eight ground robots, including dynamic objects (\eg, vehicles and pedestrians) and varying lighting conditions in environments, such as urban outdoor scenes and indoor environments like tunnels.
In addition, the SubT-MRS~\cite{subtmrs} and CERBERUS~\cite{cerberus} datasets are collected during the DARPA Subterranean Challenge, and they include various challenging cases, such as featureless surfaces and self-similar layouts.

Unlike most C-SLAM datasets that are mainly acquired in urban outdoor scenes or limited indoor environments, our \texttt{CSE} dataset covers various indoor environments specialized for service robots.
Furthermore, existing C-SLAM datasets do not reflect the characteristics that occur when real robots move since humans control the robot manually.
{In contrast}, we leverage the ROS Navigation Stack\footnote{\scriptsize\texttt{\url{https://wiki.ros.org/navigation}}} for robot driving, which {enable} more realistic scenarios, allowing the robots to recognize their environment and avoid dynamic objects.


\section{Dataset for C-SLAM in Service Environments}  \label{sec:method}

The \texttt{CSE} dataset is built upon the NVIDIA Isaac Sim~\cite{isaac_sim}. 
The NVIDIA Isaac Sim is a robotics simulator powered by the Omniverse platform that provides photo-realistic and physically accurate virtual environments.
Thus, it allows us to collect accurate GT poses and time-synchronized sensor data.
Our \texttt{CSE} dataset is acquired {using three robots} in three indoor service environments with challenging cases, including serious occlusions by dynamic objects, homogeneous floors, and redundant objects,~\etc
Each environment is categorized into static and dynamic, thereby providing 18 sequences for SLAM (6 sequences for C-SLAM).
In addition, we provide GT point clouds for each environment, which are available on the project page.
Details of our dataset, including the characteristics of scenes and scenarios, are available in the supplementary video. 

\subsection{Robot Configurations}\label{subsection:AA}
\begin{figure}[t]
    \centering
    \includegraphics[width=.95\linewidth]{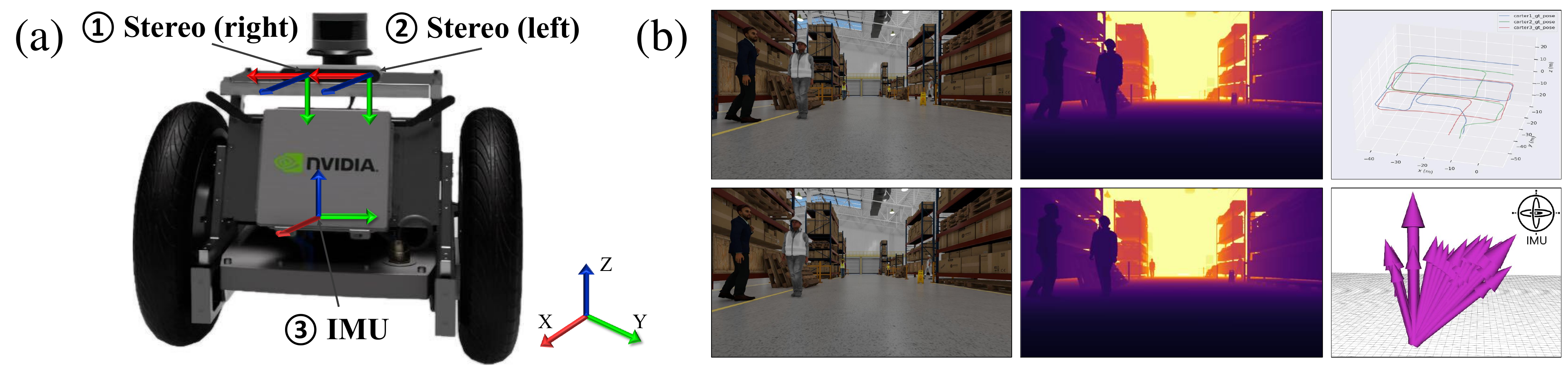}
    \caption{
        \textbf{Robot configuration and sensor data example}.
        (a) The NVIDIA Carter, our robot platform. 
        (b) Examples of acquired sensor data ({stereo RGB}, {stereo depth}, GT poses and IMU).
        }
    \label{fig:sensor_configuration}
    \vspace{-2mm}
\end{figure}

As a robot platform, we utilize the NVIDIA Carter provided by Isaac Sim (see \Fref{fig:sensor_configuration}(a)).
Carter\footnote{\scriptsize\texttt{\url{https://docs.nvidia.com/isaac/archive/2020.2/doc/tutorials/carter\_hardware.html}}\label{carter_footnote}} is a differentiable drive robot with two wheels on each side that is designed to verify the capabilities of the Isaac SDK.
Isaac SDK is intended to develop applications for complicated use cases, such as delivery robots, and the Carter is developed as a delivery robot.
Accordingly, we {choose} Carter as our robot platform for generating the dataset tailored for service robots.
We deploy three Carters as service robots in the target environments.
%
We operate this {Carter} using ROS Navigation Stack integrated within Isaac Sim.
ROS Navigation Stack is a 2D navigation stack that generates a path to a target pose using odometry and sensor data.
{It is utilized in real-world robot navigation, employing a global planner to plan the global path and a local planner to detect and avoid surrounding dynamic objects.}
By utilizing the ROS Navigation Stack, we can simulate behaviors that occur when real robots navigate.

Note that our approach reflects realistic scenarios where robots are aware of their surroundings and interact with dynamic objects.
On the other hand, existing datasets~\cite{s3e, airmuseum, graco, tian2023resilient} are generated by humans controlling the robots manually.
These approaches do not reflect the actions shown by real robots as they perceive their surroundings and encounter unexpected dynamic objects.

\subsection{Sensor Configurations}\label{subsection:BB}

For each robot, we attach several {sensor modalities} for perception in service environments, as shown in \Fref{fig:sensor_configuration}(a). 
Specifically, the sensor system of each robot is equipped with {stereo RGB/depth cameras} and an IMU sensor. 
The resolution of both stereo RGB and depth cameras is 1280$\times$720, and the baseline length is set to 12cm, following the specification\footref{carter_footnote} of the camera on board Carter.
We also provide camera parameters, such as intrinsic and extrinsic between sensors, with the left RGB camera as the reference.
For IMU, we provide empirically tuned IMU parameters.

Using this sensor system, we can acquire stereo RGB images, stereo depth images, IMU measurements, and GT poses for each robot through the ROS, as shown in \Fref{fig:sensor_configuration}(b).
In addition, we provide a function that extracts 3D sparse point clouds (\ie, pseudo-LiDAR measurements) from the depth images.
We employ this method due to constraints in extracting the 3D point cloud from the simulator.
Details of sensor specifications, ROS topics, noise levels and pseudo-LiDAR are available in our project page.

\begin{figure*}[t]
\centering
\includegraphics[width=0.85\linewidth]{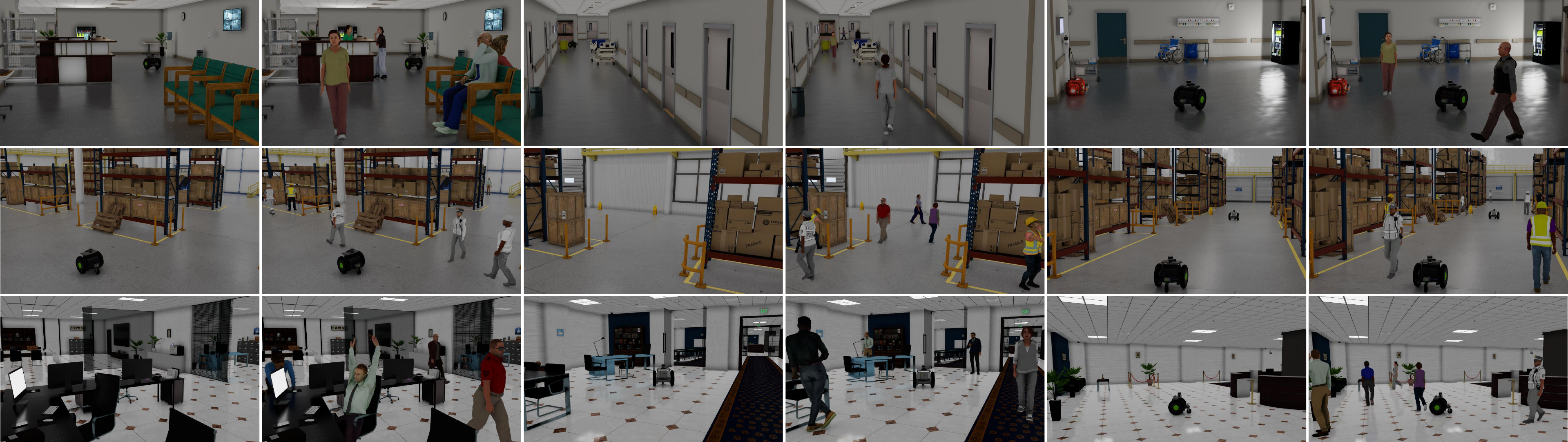} 
\caption{
\textbf{Example of service environments in the proposed \texttt{CSE} dataset}.    
        Each row shows the service environments we built (\emph{Hospital}, \emph{Warehouse}, and \emph{Office} in order) from several viewpoints. Odd columns represent static environments, while even columns represent dynamic environments.
        In particular, we can observe dynamic objects having suitable actions and clothes for each environment.
    }
    \vspace{-3mm}
\label{fig:scene_example}
\end{figure*}

\begin{figure}
    \centering
    \includegraphics[width=.85\linewidth]{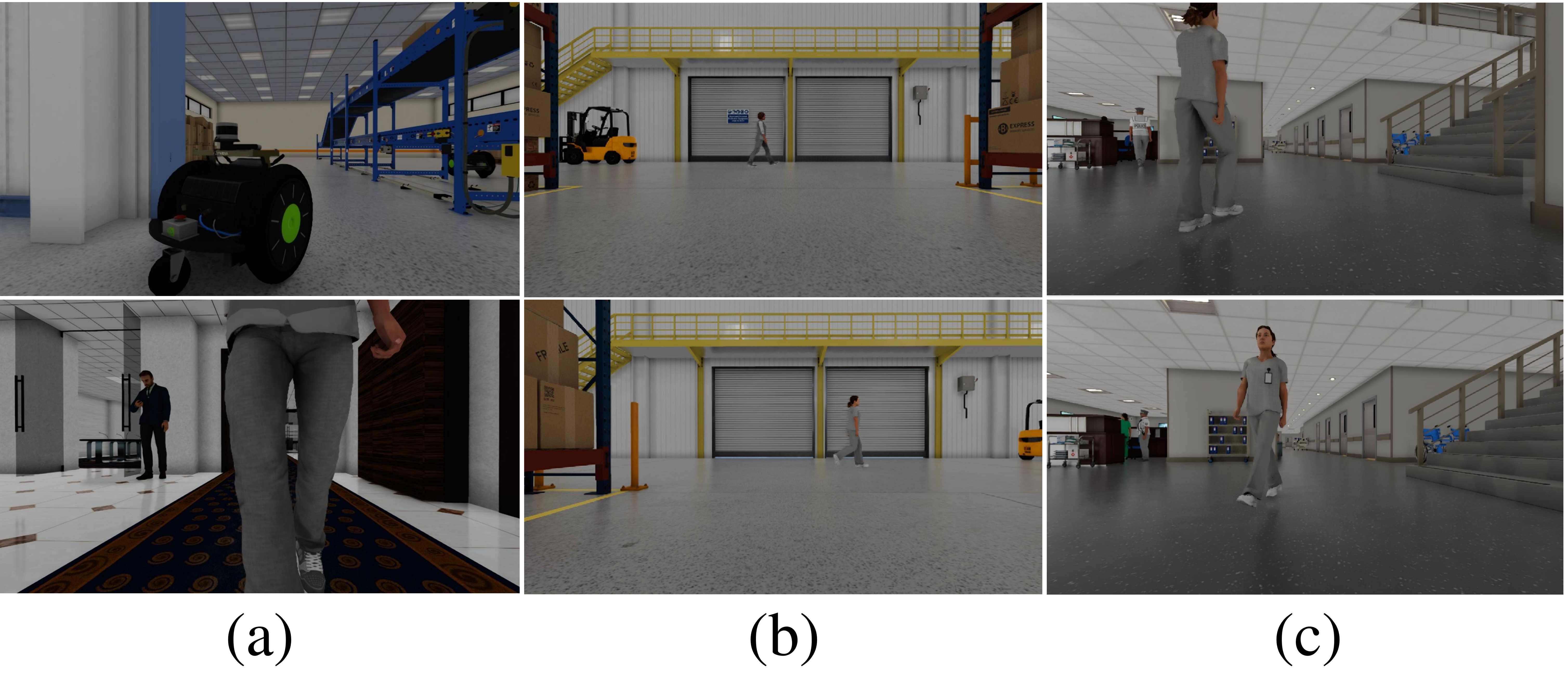}
    \\
    \caption{
        {\textbf{Challenging cases in the~\cse~dataset}. 
        (a) Occlusions from dynamic objects.
        (b) Place recognition failure due to similar structure at different location.
        (c) Invalid feature matching due to a dynamic object at different times.
        (b) and (c) are cases where SLAM failed.}
    }
     \vspace{-1mm}
    \label{fig:challenging_cases}
\end{figure}

\subsection{Service Environments}\label{subsection:CC}
We select three common indoor service environments: \emph{Hospital}, \emph{Warehouse}, and \emph{Office}, where real service robots operate.
Furthermore, we categorize each environment into static and dynamic, building a total of six environments. 
Figure~\ref{fig:scene_example} shows the examples of each environment, including static and dynamic environments.
By categorizing each environment into static and dynamic, we expect the following effects from our dataset.
1)~It {becomes} possible to generate data {with} challenging cases in the service environment itself (static) and external challenges that occur due to the addition of dynamic objects.
2)~These categorizations provide an opportunity to evaluate the effectiveness of SLAM algorithms handling dynamic objects.
3)~Moreover, the robot {navigates} with the same goal point in both static and dynamic environments, which makes the evaluation more valid.

\vspace{1mm} \noindent \textbf{Service Scenes.} \ 
We utilize basic scenes provided by NVIDIA and modify them with additional construction for our purposes.
In the case of \emph{Hospital} and \emph{Office}, we proceed with structural and textural modifications to reflect the challenging cases.
In addition, we build realistic environments by manually placing various assets suitable for each environment. 
For \emph{Warehouse}, we use the SceneBlox tool of NVIDIA Omniverse Replicator to generate a basic scene.
After building the basic scene, we do the same post-processing as \emph{Hospital} and \emph{Office}.

\emph{Hospital}, with 76m$\times$45m, mainly consists of narrow and long corridors. 
Specifically, {the} walls and floors of \emph{Hospital} are homogeneous, making it hard to extract the visual information and re-localization difficult. 
{For example}, there are a series of doors and objects with the same design in the corridors.
In other words, there are lots of redundant objects placed, which can create ambiguity in feature matching and significantly reduce pose estimation accuracy.
\emph{Office}, with 31m$\times$94m, comprises two large spaces connected by short corridors.
It {includes} the various challenging cases since it has homogeneous walls/floors and ceilings with repetitive patterns.
The floor is made up of reflective materials; these reflective properties, especially, can cause another challenge by light reflections.
\emph{Warehouse} is a cuboid shape with 56m$\times$74m, where large grid structures are regularly arranged.
Similar to \emph{Hospital}, it consists of homogeneous walls and floors, as well as a large number of redundant doors and boxes, which can make feature matching challenging.

\vspace{1mm} \noindent \textbf{Dynamic  Objects.} \ 
All dynamic environments are built by placing dynamic objects based on each static environment.
As dynamic objects, we utilize human assets provided by NVIDIA and purchased assets from ActorCore\footnote{\scriptsize\texttt{\url{https://actorcore.reallusion.com/}}}.
~Dynamic objects are evenly distributed spatially and perform suitable actions {to} each environment.
For example, in the \emph{Hospital}, there are doctors walking around the rooms or {nurses} talking at the reception desk.
For \emph{Office}, as shown in~\Fref{fig:scene_example}, the office workers are either sitting in chairs or talking to each other.
In addition, dynamic objects move around specific areas continuously.
Through these motions,
severe occlusions can occur, obstructing the camera view (see \Fref{fig:challenging_cases}(a)).

\begin{figure*}[t]
	\centering
	\includegraphics[width=0.9\linewidth]{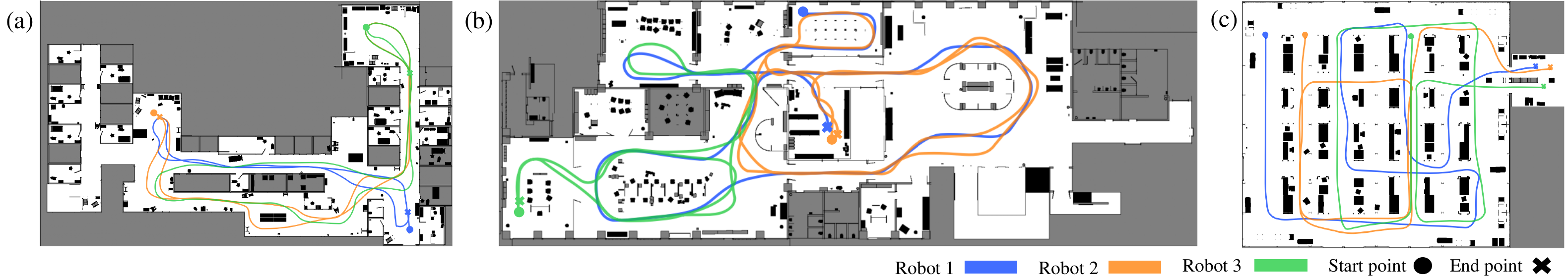}
	\caption{
        \textbf{Scenarios in the proposed \texttt{CSE} dataset}.
            We visualize scenarios for each dynamic environment on its 2D occupancy map with the same scale. 
            Note that scenarios in this illustration only show dynamic environments. 
        }
        \vspace{-3mm}
	\label{fig:scenarios}
\end{figure*}

\subsection{Scenarios}\label{subsection:DD}

C-SLAM involves multiple robots collaborating to explore the environments and build a map. 
During this process, each robot visits the same location repeatedly (\ie, intra-robot loop closure) and frequently encounters each other (\ie, inter-robot loop closure). 
Such intra-/inter-robot loop closures are essential for the robots to estimate their pose and build the map accurately.
Based on this fact, we design the scenarios considering the various interactions, including intra-/inter-robot loop closures.

{Specifically, the interactions we considered are categorized into three types: Intra-robot, Robot-to-robot, and Robot-to-human interactions. 
Firstly, the intra-robot interaction refers to all cases where intra-robot loop closure occurs. 
Secondly, robot-to-robot interaction is defined by all phenomena that lead to inter-robot loop closure through the exchange of data between robots. We categorize this case into three main scenarios (\textsf{\small Follow}, \textsf{\small Intersection}, and \textsf{\small Revisit}).
\textsf{\small Follow} describes a situation where one robot follows behind another, and \textsf{\small Intersection} means different robots crossing each other while facing one another simultaneously.
Finally, \textsf{\small Revisit} refers to a situation where a robot passes through the same path that another robot passed earlier, but at a different time.
Details of each type are available in the supplementary video. 
{Lastly, our dataset includes cases where a human recognizes and avoids a robot at close range, which we define as robot-to-human interaction.}
In summary, we design the scenarios considering these various interactions.
However, our dataset is generated under the assumption that the given indoor environments would have constant communication. Therefore, we do not consider the concept of bandwidth or links between the robots.
}

For each environment, the static and dynamic scenarios share the same goal points for robot navigation. 
However, due to variability introduced by the ROS Navigation Stack, the paths between goal points in static and dynamic scenarios are not perfectly identical but are almost similar. 
Additionally, across all environments, the dynamic scenarios share the common feature of people avoiding the robot at close range, leading to challenges such as occluding the view of the robot, as illustrated in~\Fref{fig:challenging_cases}(a).
For this reason, the following part of this section focuses on explaining the static scenarios. 
It should be noted that the dynamic scenario includes all features of the static scenario in each environment.
The robot scenarios for each environment can be found in~\Fref{fig:scenarios}, and their characteristics are detailed in~\Tref{method:scenarios}.

\vspace{1mm} \noindent \textbf{Hospital.} \ 
The scenarios for \emph{Hospital} are depicted in \Fref{fig:scenarios}(a).
All robots start from different starting points and complete their navigation by returning to their starting points.

The scenarios for all robots consist of driving down long corridors and exploring small spaces. 
In common, they perform large rotations when exploring small spaces to avoid collision.
These large rotations can quickly change the view of the camera, making pose estimation difficult. 
\textsc{Robot 1} and \textsc{Robot 3} perform intra-robot loop closures when they reach the endpoints, while \textsc{Robot 2} does not. 
Instead, \textsc{Robot 2} explores the hospital as a whole, resulting in frequent \textsf{\small Revisit}, which lead to numerous inter-robot loop closures.
In addition, \textsc{Robot 3} encounters other robots (\textsf{\small Intersection}) and follows \textsc{Robot 1} (\textsf{\small Follow}), both leading to inter-robot loop closures.

\vspace{1mm} \noindent \textbf{Office.} \ 
Figure~\ref{fig:scenarios}(b) is the scenarios for \emph{Office}.
Similar to scenarios in \emph{Hospital}, every robot has different starting points.
\textsc{Robot 2} and \textsc{Robot 3} finish their navigation near their starting points, whereas \textsc{Robot 1} ends at the different location from its starting point.

\textsc{Robot 2} and \textsc{Robot 3} drive over a specific region repeatedly, and \textsc{Robot 1} drives {throughout} the overall area of office.
In the case of \textsc{Robot 2}, it consists of complicated scenarios that involve driving through both rooms and corridors. 
This causes \textsc{Robot 2} to make large rotations when going through the narrow doors.
The scenarios for \textsc{Robot 2} and \textsc{Robot 3} have many intra-robot loop closures due to repeatedly driving over the same areas.
However, in the case of \textsc{Robot 1}, there are no intra-robot loop closures. Instead, it has lots of \textsf{\small Revisit} that leads the inter-robot loop closures because it drives through the entire environment.

\vspace{1mm} \noindent \textbf{Warehouse.} \ 
Figure~\ref{fig:scenarios}(c) shows \warehouse~scenarios, where all robots start from different locations and complete their navigation at the same space.

In \warehouse, the robots drive between large, regularly aligned structures.
The scenarios for \textsc{Robot 2} and \textsc{Robot 3} involve intra-robot loop closures, while \textsc{Robot 1} does not contain any.
However, for \warehouse~scenarios, all robots contain many inter-robot loop closures.
For example, \textsc{Robot 1} drive back through the area that \textsc{Robot 3} passed through (\textsf{\small Revisit}), and \textsc{Robot 2} drive behind \textsc{Robot 1}~(\textsf{\small Follow}). 
In particular, \textsc{Robot 1} and \textsc{Robot 2} intersect each other, which {leads to} the challenges of observing one another at close range, resulting in extreme occlusion.
In addition, \textsc{Robot 1} and \textsc{Robot 2} overlap their driving paths in narrow sections, causing \textsc{Robot 2} to perform a recovery behavior.
Recovery behavior means the robot independently recognizes and solves unexpected obstacles and dynamic changes during path planning to continue driving.
Specifically, \textsc{Robot 2} sees \textsc{Robot 1} as a dynamic obstacle and stops, then resumes driving once it is sure that it is no longer obstructing its path.
These scenarios well reflect real-world situations encountered by service robots.

\begin{table*}[]
    \centering
    \caption{\textbf{Dynamic scenario configurations.}}
    \small
    \label{method:scenarios}
    \begin{threeparttable}
    \resizebox{0.85\linewidth}{!}{
    \begin{tabular}{c|c|c|c|c|ccc|c|c|c}
    \toprule
            \multirow{3}{*}{Environment} & \multirow{3}{*}{Robot} & \multirow{3}{*}{Duration (s)} & \multirow{3}{*}{Length (m)} & \multicolumn{4}{c|}{Loop Closure} 
            & \multirow{3}{*}{Characteristics / Challenges$^\dagger$} & \multirow{3}{*}{Size} & \multicolumn{1}{c}{\multirow{1}{*}{\# of}} \\ \cline{5-8} 
            & & & & \multirow{2}{*}{Intra} & \multicolumn{3}{c|}{Inter} & & & \multicolumn{1}{c}{\multirow{1}{*}{dynamic}}  \\
            & & & & & \textsf{\small Follow} & \textsf{\small {Intersection}} & \textsf{\small Revisit} & &  & \multicolumn{1}{c}{\multirow{1}{*}{objects\tnote{*}}}  \\

            \midrule
            \multirow{3}{*}{\emph{Hospital}} & \multirow{1}{*}{\textsc{Robot 1}} & $373.1$ & $124.0$ & \checkmark &  & \checkmark & \checkmark & \multirow{3}{*}{Long corridors / Homogeneous floor} & \multirow{3}{*}{$76$m$\times$$45$m} & \multirow{3}{*}{$3+16$}  \\
             & \multirow{1}{*}{\textsc{Robot 2}} & $563.1$ & $200.6$ &  &  & \checkmark & \checkmark &  &  &    \\            
             & \multirow{1}{*}{\textsc{Robot 3}} & $509.1$ & $182.8$ & \checkmark & \checkmark & \checkmark & \checkmark &  &  &    \\   \cmidrule{1-11}       

            \multirow{3}{*}{\emph{Office}} & \multirow{1}{*}{\textsc{Robot 1}} & $604.5$ & $210.1$ &  & \checkmark &  & \checkmark & \multirow{1}{*}{Complex space with rooms and corridors /} & \multirow{3}{*}{$31$m$\times$$94$m} & \multirow{3}{*}{$3+37$}  \\
             & \multirow{1}{*}{\textsc{Robot 2}} & $703.5$ & $238.2$ & \checkmark & \checkmark &  & \checkmark & \multirow{1}{*}{Repetitive pattern (floor, ceiling), } &  &    \\            
             & \multirow{1}{*}{\textsc{Robot 3}} & $508.5$ & $175.9$ & \checkmark &  &  & \checkmark & \multirow{1}{*}{Reflective material floor} &  &    \\      \cmidrule{1-11} 

                         \multirow{3}{*}{\emph{Warehouse}} & \multirow{1}{*}{\textsc{Robot 1}}  & $645.0$ & $250.1$ &  & \checkmark & \checkmark & \checkmark & \multirow{3}{*}{Large regular grid structures / Homogeneous floor} & \multirow{3}{*}{$56$m$\times$$74$m} & \multirow{3}{*}{$3+38$}  \\
             & \multirow{1}{*}{\textsc{Robot 2}} & $643.1$ & $254.7$ & \checkmark &  & \checkmark & \checkmark &  &  &    \\            
             & \multirow{1}{*}{\textsc{Robot 3}} & $631.1$ & $251.0$ & \checkmark & \checkmark & \checkmark & \checkmark &  &  &    \\            
             \bottomrule
        \end{tabular}
       }
       \begin{tablenotes}
            \item \scriptsize{$\dagger$ In common, service environments in our dataset include visual redundancy, dynamic objects, and homogeneous walls.}
            \item \scriptsize{* The number of objects that include robots and humans.}
       \end{tablenotes}
    \end{threeparttable}
    \vspace{-3mm}
\end{table*}


\section{Experiments}
In this section, we describe SLAM algorithms utilized to evaluate our dataset. 
We also provide the experimental analysis and visualization results of various SLAM baselines~(see \emph{Hospital} in~\Fref{fig:experiments}).
Visualization results for all environments are available in the supplementary video.

\subsection{Baseline}

\noindent \textbf{SLAM for Single-Robot.} \
To evaluate our dataset, we utilize ORB-SLAM3~\cite{orb3} and VINS-Fusion~\cite{vins-fusion} as baselines for single-robot SLAM.
ORB-SLAM3 facilitates a range of camera configurations and demonstrates outstanding performance by leveraging pre-constructed maps in scenarios with restricted visual data.
VINS-Fusion is an optimization-based odometry framework that utilizes visual and inertial information, integrating sensor data into pose graph optimization for accurate position estimation.

\vspace{1mm} \noindent \textbf{C-SLAM for Multi-Robot.} \
Unlike SLAM for single-robot, C-SLAM for multi-robot performs SLAM tasks using multiple robots.
COVINS~\cite{covins} is a centralized visual-inertial SLAM system that utilizes data collected from multiple robots.
This system gathers data generated by ORB-SLAM3 from numerous robots to perform global optimization, and it enhances joint estimation by incorporating place recognition and eliminating redundant data.
Swarm-SLAM~\cite{swarm} employs a decentralized approach, utilizing novel techniques for efficient communication between robots and rapid convergence. 
This framework is well-suited for large-scale deployment, and its effectiveness in terms of accuracy and resource utilization efficiency has been demonstrated through empirical experiments.

{Additionally, to evaluate the impact of dynamic objects, we develop Swarm-SLAM-D and COVINS-D by applying the dynamic feature removal module that handles dynamic objects into Swarm-SLAM and COVINS.
This module is modified based on the moving consistency check module of DS-SLAM~\cite{dsslam}, and identifies and removes dynamic features using both previous and current frames.
Detailed explanation and code are available on the project page.}

\subsection{Experimental Setup}
To evaluate various sensor modalities offered by our dataset, we conduct evaluations using RGB-D, stereo-inertial, and mono-inertial modalities provided by each baseline algorithm.
The trajectories from each baseline are measured for performance using the absolute trajectory error (ATE).
The all evaluation is conducted with EVO~\cite{evo}.

For single-robot SLAM algorithms, we set the play rate to $1.0\times$ and evaluate all sequences. For multi-robot SLAM algorithms, we use the $0.5\times$ play rate and evaluate with three robots concurrently. This adjustment ensures smooth running, given the high computational demands of multi-robot SLAM.

\subsection{Results and Analysis}

\noindent \textbf{SLAM for Single-Robot.}  \ 
In \Tref{exp:single_slam}, the evaluation results using single-robot SLAM are presented. 
We observe that both ORB-SLAM3 and VINS-Fusion achieve high accuracy with the stereo-inertial setup on average. 
Notably, in \emph{Hospital}, the presence of numerous static objects, such as chairs and desks, facilitates easier feature matching, thereby enhancing the accuracy of pose estimation.
However, due to the challenging cases of each environment, specific sequences lead to poor estimation results.
For example, we observe that the \textsc{Robot 3} performs incorrect place recognition in ORB-SLAM3.
This occurs due to redundant objects, as shown in \Fref{fig:challenging_cases}(b), resulting in low accuracy.
{Additionally, ORB-SLAM3 sometimes halts during operation.}
{Specifically, in certain scenarios, relatively small IMU values occur, at which time the IMU processing logic of ORB-SLAM3 reacts sensitively, leading to algorithmic failures.}
{
For VINS-Fusion, we observe that inadequate feature matching occurs in \warehouse~when robots encounter each other at close range, as the other robot is treated as a dynamic object.
Through this result, we believe that by exchanging their positions, robots could better handle dynamic objects (\ie, other robots), improving performance.
Furthermore, we can see that these challenges adversely affect the accuracy of SLAM, highlighting the need to address dynamic objects and environmental complexities to enhance overall performance.
}

\vspace{1mm} \noindent \textbf{C-SLAM for Multi-Robot.} \ 
\Tref{exp:c_slam} shows the evaluation results of multi-robot SLAM on our dataset. 
In static environments, Swarm-SLAM shows robust performance compared to other multi-robot SLAM algorithms {on RGB-D and stereo setups}. 
In contrast, in dynamic environments, there is a notable reduction in performance, particularly in \emph{Hospital}. 
{However, Swarm-SLAM-D shows higher accuracy compared to Swarm-SLAM due to the dynamic feature removal module.
It indicates that the dynamic objects have a significant impact on the performance of Swarm-SLAM. 
However, the results of Swarm-SLAM-D degrade for specific sequences.
This happens because the naive dynamic feature removal approach may result in improper feature removal in specific sequences.
}

For COVINS, it shows inferior performance and numerous failures throughout our dataset. 
{This is because centralized SLAM shares a single server (\ie, backend) among all robots; a failure case occurring in a single robot can also affect another.}
{Additionally, COVINS sometimes halts during operation when relatively small IMU values occur due to similar issues to those in ORB-SLAM3.}
{
However, for COVINS-D, we observe significant performance improvements in specific sequences, such as Robot 2 in the dynamic \warehouse, by handling dynamic objects.
On the other hand, we also observe a performance degradation compared to COVINS in specific sequences.
{
This issue arises when extreme occlusion from dynamic objects occurs in the robot view, causing the dynamic feature removal module to remove an excessive number of keypoints. 
In such cases, the RANSAC-based dynamic feature removal module is affected by the number of features, leading to misclassification of static and dynamic keypoints, which negatively impacts performance.
}
Based on these results, we believe that utilizing the information from dynamic objects (\eg, dynamic object tracking) may be more effective than simply removing them.}

In addition, we observe that multi-robot SLAM can fail even if the place recognition module performs well. 
For example, in Swarm-SLAM, although a robot recognizes the same static environment by the place recognition module (see \Fref{fig:challenging_cases}(c)), dynamic objects observed in the viewpoint can cause invalid feature matching.
This can lead to inaccurate bundle adjustment.
This rare edge case in single-robot SLAM becomes more common in multi-robot SLAM, where robots exchange observations in various encounter cases.
{Moreover, we observe that when there are lots of overlapping regions between robots, it leads to excessive inter-robot loop closures, which significantly increase optimization complexity and processing time.
We believe this issue is inevitable for service robots continuously operating in the same space, and more efficient backend solutions are needed.
}

In summary, the proposed \texttt{CSE} dataset can help analyze how the characteristics contained in the service environments affect SLAM algorithms and have the potential to improve SLAM performance in the service environments. 

\begin{table}[t]
    \centering
    \caption{\textbf{{Evaluation of SLAM for single-robot.}}}
    \small
    \label{exp:single_slam}
    \begin{threeparttable}
    \resizebox{0.99\linewidth}{!}{
    \begin{tabular}{c|c|c|>{\centering\arraybackslash}m{1.3cm}|>{\centering\arraybackslash}m{1.3cm}|>{\centering\arraybackslash}m{1.3cm}|>{\centering\arraybackslash}m{1.3cm}|>{\centering\arraybackslash}m{1.3cm}}
    \toprule 
               &  &  &\multicolumn{3}{c|}{ORB-SLAM3~\cite{orb3}}   & \multicolumn{2}{c}{VINS-Fusion~\cite{vins-fusion}}   \\
            \midrule 
            Sequences & & & \multicolumn{1}{c|}{RGB-D} & \multicolumn{1}{c|}{\makecell{Mono\\Inertial}} & \multicolumn{1}{c|}{\makecell{Stereo\\Inertial}} & \multicolumn{1}{c|}{\makecell{Mono\\Inertial}} & \multicolumn{1}{c}{\makecell{Stereo\\Inertial}} \\
            \midrule 
             \multirow{9}{*}{Static} & \multirow{3}{*}{\emph{Hospital}} & \textsc{R1} & $\textbf{0.015}$ & $0.191$ & $0.017$ & $4.863$ & $0.331$ \\
             &  & \textsc{R2} & $\textbf{0.057}$ & $16.056$ & $0.061$ & $0.513$  & $0.210$  \\
             &  & \textsc{R3} & $\textbf{0.040}$ & $7.897$ &  \xmark & $2.555$ & $0.387$  \\  \cmidrule{2-8}
              & \multirow{3}{*}{\emph{Office}} & \textsc{R1} & $0.058$ & $3.764$  & $\textbf{0.038}$ & $2.042$ & $0.213$ \\
             &  & \textsc{R2} & $\textbf{0.023}$ & $8.903$ & $6.676$ & $3.513$ & $0.471$ \\
             &  & \textsc{R3} & $3.432$ & $\textbf{0.094}$ & $3.040$ & $4.244$ & $0.170$ \\ \cmidrule{2-8}
              & \multirow{3}{*}{\emph{Warehouse}} & \textsc{R1} & $\textbf{0.565}$ & \xmark & \xmark & $17.027$ & $1.809$ \\
             &  & \textsc{R2} & $\textbf{0.030}$ & $0.198$ & $\textbf{0.030}$ & $10.726$ & $0.328$ \\
             &  & \textsc{R3} & $16.490$ & $\textbf{0.373}$ & $4.707$ & $5.832$ & $1.957$ \\ \midrule 
             \multirow{9}{*}{Dynamic} & \multirow{3}{*}{\emph{Hospital}} & \textsc{R1} & $0.023$ & $0.108$ & $\textbf{0.019}$ & $2.244$ & $0.525$ \\
             &  & \textsc{R2} & $\textbf{0.098}$ & \xmark & \xmark & $1.999$ & $0.346$ \\
             &  & \textsc{R3} & $\textbf{0.030}$ & $10.340$ & $0.033$ & $3.376$ & $0.348$ \\ \cmidrule{2-8}
              & \multirow{3}{*}{\emph{Office}} & \textsc{R1} & $\textbf{0.090}$ & $0.256$ & $13.424$ & $2.546$ & $0.253$ \\
             &  & \textsc{R2} & $\textbf{0.060}$ & \xmark & \xmark & $6.740$ & $0.442$ \\
             &  & \textsc{R3} & $0.067$ & $0.072$ & $\textbf{0.022}$ & $3.450$ & $0.263$ \\ \cmidrule{2-8}
              & \multirow{3}{*}{\emph{Warehouse}} & \textsc{R1} & $\textbf{0.054}$ & $0.552$ & $0.068$ & $4.322$ & $1.322$ \\
             &  & \textsc{R2} & $\textbf{0.021}$ & $0.259$ & $0.035$ & $4.268$ & $0.496$ \\
             &  & \textsc{R3} & $16.682$ & $0.399$ & $\textbf{0.076}$ & $5.385$ & $0.364$ \\ 
            \bottomrule
         \end{tabular}
       }
       \begin{tablenotes} 
            \item \scriptsize{\xmark \, Fail to obtain trajectory due to algorithm halt during operation.}
            \item \scriptsize{R1-3 \,indicate the robot names~(\eg, R1 means Robot 1).}
            \item \scriptsize{{Note that the best result for each scenario are bolded.}}
            \item The metric represents RMS ATE in meters.
       \end{tablenotes}
       \vspace{-3mm}
    `    \end{threeparttable}
\end{table}

\begin{table}[t]
    \centering
    \caption{\textbf{{Evaluation of C-SLAM for multi-robot}.}}
    \label{exp:c_slam}
    \begin{threeparttable}
    \resizebox{1.0\linewidth}{!}{%
    \begin{tabular}{c|c|c|>{\centering\arraybackslash}m{1.2cm}|>{\centering\arraybackslash}m{1.2cm}|>{\centering\arraybackslash}m{1.2cm}|>{\centering\arraybackslash}m{1.2cm}|>{\centering\arraybackslash}m{1.2cm}|>{\centering\arraybackslash}m{1.2cm}}
    \toprule 
               &  &  & \multicolumn{1}{c|}{\makecell{COVINS\\~\cite{covins}}}  & \multicolumn{1}{c|}{\makecell{COVINS-D \\ (w/ Dynamic)}} & \multicolumn{2}{c|}{Swarm-SLAM~\cite{swarm}} & \multicolumn{2}{c}{\makecell{Swarm-SLAM-D \\ (w/ Dynamic)}}  \\
            \midrule 
             Sequences & & & \multicolumn{1}{c|}{\makecell{Mono\\Inertial}} & \multicolumn{1}{c|}{\centering {\makecell{Mono\\Inertial}}} & \multicolumn{1}{c|}{RGB-D} & \multicolumn{1}{c|}{Stereo} & \multicolumn{1}{c|}{RGB-D} & \multicolumn{1}{c}{Stereo} \\
            \midrule 
             \multirow{9}{*}{Static} & \multirow{3}{*}{\emph{Hospital}} & \textsc{R1} & $9.431$ & \centering $12.959$ & $\textbf{
             0.176}$ & $0.387$ &{0.213} & {0.312} \\
             &  & \textsc{R2} & $2.010$ & \centering $1.346$ & $\textbf{0.149}$ & $0.405$ & {0.339} & {0.445} \\
             &  & \textsc{R3} & $\triangle$ & \centering $\triangle$ & $\textbf{0.130}$ & $0.318$ & {0.298} & {0.404} \\ \cmidrule{2-9}
              & \multirow{3}{*}{\emph{Office}} & \textsc{R1} & $9.232$ & \centering $10.921$ & $1.281$ & $\textbf{0.104}$ & {0.157} & {0.107} \\
             &  & \textsc{R2} & $8.917$ & \centering $9.1428$ & $1.288$ & $\textbf{0.121}$ & {1.517} & {2.030}\\
             &  & \textsc{R3} & $8.774$ & \centering $1.129$ & $0.739$ & $\textbf{0.067}$ & {1.704} & {0.198} \\ \cmidrule{2-9}
              & \multirow{3}{*}{\emph{Warehouse}} & \textsc{R1} & \xmark & \centering \xmark & $0.827$ & $\textbf{0.189}$ & {3.326} & {0.397} \\
             &  & \textsc{R2} & \xmark & \centering \xmark & $0.697$ & $0.165$ & {1.309} & \textbf{0.140} \\
             &  & \textsc{R3} & \xmark & \centering \xmark & $0.588$ & $\textbf{0.120}$ & {2.336} & {0.217} \\ \midrule
             \multirow{9}{*}{Dynamic} & \multirow{3}{*}{\emph{Hospital}} & \textsc{R1} & \xmark & \centering \xmark & $9.441$ & $13.670$ & $0.585$ & $\textbf{0.323}$ \\
             &  & \textsc{R2} & \xmark & \centering \xmark & $1.797$ & $\textbf{0.426}$ &  $1.586$ & $0.440$ \\
             &  & \textsc{R3} & \xmark & \centering \xmark & $8.431$ & $1.522$ & $2.343$ & $\textbf{0.739}$  \\ \cmidrule{2-9}
              & \multirow{3}{*}{\emph{Office}} & \textsc{R1} & \xmark & \centering \xmark & $0.888$ & $0.113$ & $0.198$ & $\textbf{0.089}$ \\
             &  & \textsc{R2} & \xmark & \centering \xmark & $0.721$ & $0.111$ & $0.376$ & $\textbf{0.072}$ \\
             &  & \textsc{R3} & \xmark & \centering \xmark & $0.491$ & $\textbf{0.079}$ & $0.229$ & $0.126$ \\ \cmidrule{2-9}
              & \multirow{3}{*}{\emph{Warehouse}} & \textsc{R1} & $9.560$ & \centering $8.252$ & $0.894$ & $\textbf{0.228}$ & $15.721$ & $0.313$ \\
             &  & \textsc{R2} & $7.203$ & \centering $2.176$ & $0.582$ & $0.192$ & $0.231$ & $\textbf{0.159}$ \\
             &  & \textsc{R3} & $11.539$ & \centering $13.168$ & $0.670$ & $0.151$ & $15.031$ & $\textbf{0.122}$ \\ 
            \bottomrule
        \end{tabular}
        }
       \begin{tablenotes}
            \item \scriptsize{$\triangle$ Only \textsc{robot 3} fail due to the algorithm's shutdown.}
       \end{tablenotes}
       \vspace{-0mm}
    \end{threeparttable}
\end{table}

\begin{figure*}[]
	\centering
        \includegraphics[width=0.75\linewidth]{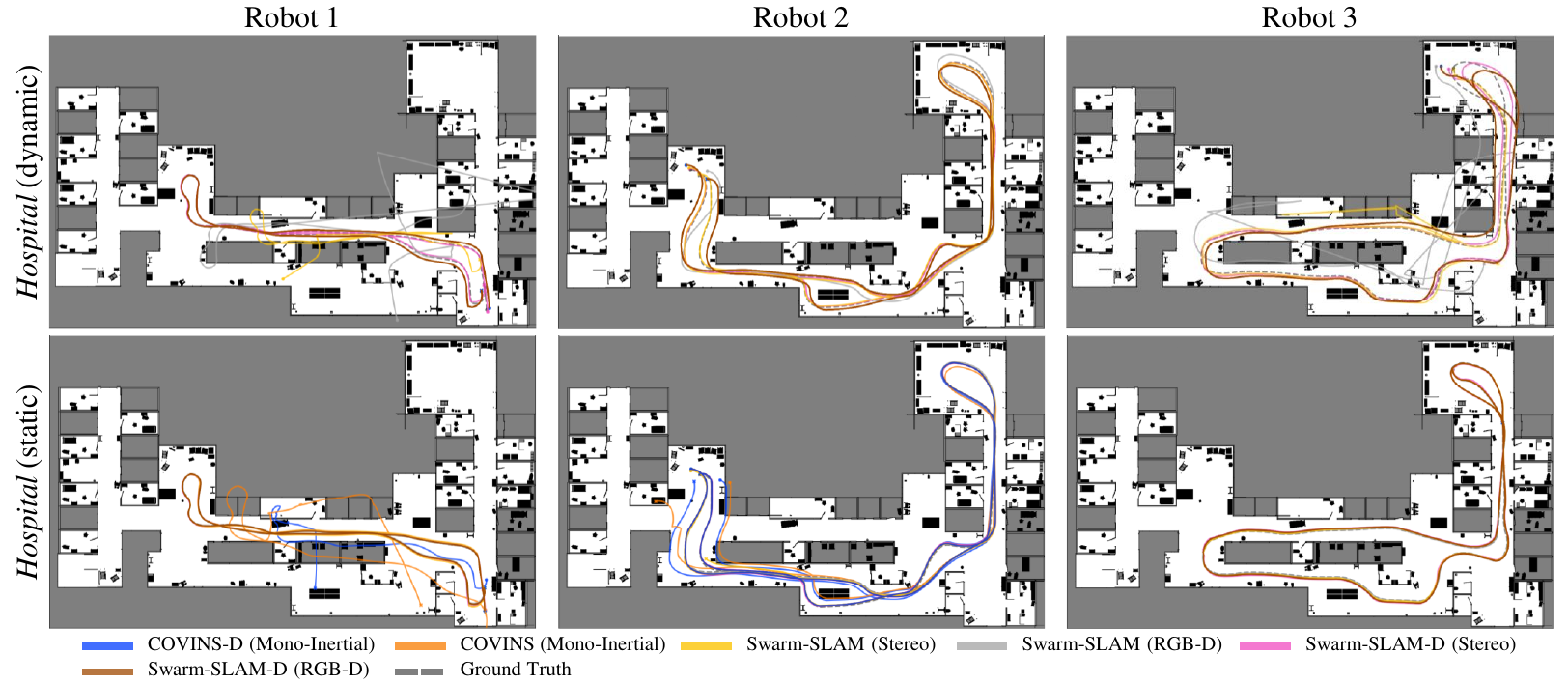}
	\caption{
	    \textbf{{The visualization results of SLAM algorithms in our dataset}}.
            {Note that we only visualize \hospital~experiments that success in full sequences. Results from other environments are available in the supplementary video.
            }
        }
        \vspace{-3mm}
	\label{fig:experiments}
\end{figure*}

\section{DISCUSSION}
In this section, we discuss the limitations of our dataset, particularly the gap between simulation and the real world. 
We also highlight the potential for enhancing SLAM research using the \cse~dataset and propose our future work. 

\subsection{Limitation}
The gap between simulation and the real world exists due to complex factors, including lighting and sensor noise.
In response, prior studies~\cite{tartanair, domainrandom} have focused on reducing the sim-to-real gap by increasing the diversity of the environment.
On the other hand, our work focuses on building a synthetic C-SLAM dataset that is difficult to obtain in real service environments due to various issues, such as handling dynamic objects and obtaining precise time-synchronized sensor data between multi-robot, rather than focusing on reducing this gap.
For example, we can reproduce diverse scenarios that robots may experience in real service environments (\eg, facing unexpected dynamic objects), and provide precisely time-synchronized sensor data. 
In addition, we confirmed that the various challenging cases (\eg, dynamic objects, visual redundancy) included in the \cse~dataset influence the performance of SLAM algorithms.
As a result, we believe that the \cse~dataset includes various challenges that SLAM still needs to solve, and can be a valuable resource for SLAM research. 
Moreover, the \cse~dataset can be realistically modeled with factors such as sensor noise, motion blur, and contrast changes through post-processing methods~\cite{Simkinect, slam_under_perturbation}.
We also provide a method that allows noise to be added to the depth image and the maximum depth range to be adjusted based on~\cite{Simkinect}. This code is available on our project page.
Through these aspects, we believe that the \cse~dataset can contribute to the development of C-SLAM research.

\subsection{Future Work}
In real service environments, various interactions cause frequent temporal scene changes~\cite{rio}.
Therefore, C-SLAM also needs to be equipped with long-term scene understanding capabilities to effectively handle such changes.
With this in mind, we plan to provide data for life-long SLAM that includes scene changes and diverse robot trajectories. 
We have already built sample data considering scene changes in \office~environment, and it is available on the project page. 
Additionally, we also plan to offer GT labels (\eg, semantic segmentation, dynamic object poses) as well as additional sensor modalities (\eg, fisheye camera) to support the development and use of various algorithms.

\section{CONCLUSION}

In this work, we propose the \texttt{CSE} dataset, a new synthetic C-SLAM benchmark dataset for multiple service robots. 
Unlike previous C-SLAM datasets that are mainly acquired in urban outdoor scenes or limited indoor environments (\eg, corridors or room-size {laboratories}), we focused on acquiring C-SLAM dataset with three robots from common indoor service environments that reflect diverse, challenging cases that may occur in real-world service environments.
Each environment is divided into static and dynamic with dynamic objects, and each robot drives through the same scenario in both environments.
This design strategy provides an opportunity to evaluate the effectiveness of single-robot SLAM and multi-robot SLAM in dealing with a variety of dynamic objects.
We also generate data that reflects the driving properties of real-world service robots since the robots recognize their surroundings and drive themselves in the simulation.
Through all these various characteristics, we expect our dataset will contribute to the advancement of C-SLAM research for multiple service robots.

\bibliographystyle{IEEEtran}
\bibliography{egbib}

\end{document}